\documentclass[10pt, a4paper]{article}
\usepackage{lrec2022} %
\usepackage{multibib}
\newcites{languageresource}{Language Resources}
\usepackage{graphicx}
\usepackage{tabularx}
\usepackage{soul}
\usepackage{titlesec}
\titleformat{\section}{\normalfont\large\bfseries\center}{\thesection.}{1em}{}
\titleformat{\subsection}{\normalfont\SmallTitleFont\bfseries\raggedright}{\thesubsection.}{1em}{}
\titleformat{\subsubsection}{\normalfont\normalsize\bfseries\raggedright}{\thesubsubsection.}{1em}{}
\renewcommand\thesection{\arabic{section}}
\renewcommand\thesubsection{\thesection.\arabic{subsection}}
\renewcommand\thesubsubsection{\thesubsection.\arabic{subsubsection}}

\usepackage{epstopdf}
\usepackage[utf8]{inputenc}

\usepackage{hyperref}
\usepackage{xstring}

\usepackage{color}

\usepackage{booktabs}
\usepackage{enumitem}

\usepackage{xstring}

\title{Comparing Formulaic Language in Human and Machine Translation: Insight from a Parliamentary Corpus}

\name{Yves Bestgen} 

\address{Laboratoire d'analyse statistique des textes\\
  Universit\'e catholique de Louvain \\
  Place Cardinal Mercier, 10 1348 Louvain-la-Neuve, Belgium \\
         yves.bestgen@uclouvain.be\\}

\abstract{
A recent study has shown that, compared to human translations, neural machine translations contain more strongly-associated formulaic sequences made of relatively high-frequency words, but far less strongly-associated formulaic sequences made of relatively rare words. These results were obtained on the basis of translations of quality newspaper articles in which human translations can be thought to be not very literal. The present study attempts to replicate this research using a parliamentary corpus. The text were translated from French to English by three well-known neural machine translation systems: DeepL, Google Translate and Microsoft Translator. The results confirm the observations on the news corpus, but the differences are less strong. They suggest that the use of text genres that usually result in more literal translations, such as parliamentary corpora, might be preferable when comparing human and machine translations. Regarding the differences between the three neural machine systems, it appears that Google translations contain fewer highly collocational bigrams, identified by the CollGram technique, than Deepl and Microsoft translations.
\\ \newline }

\begin{document}

\maketitleabstract

\section{Introduction}

Due to the success of neural machine translation systems, more and more research is being conducted to compare their translations to human translations. Most of these studies take a global view of quality \cite{Popel2020,laubli2018,Google}, but others focus on much more specific dimensions that could be further improved, such as lexical diversity and textual cohesion \cite{declercq2021,vanmassenhove2019}. 

In a recent study, \newcite{TRITON2021} analyzed the frequency of use of a specific category of formulaic sequences, the "habitually occurring lexical combinations" \cite{Laufer11}, which are statistically typical of the language because they are observed "with markedly high frequency, relative to the component words" \cite{BAL10}. To identify them, he used the CollGram technique which relies on two lexical association indices: mutual information (MI) and t-score, calculated on the basis of the frequencies in a reference corpus \cite{Berna07,BG14,Dur09}. A discussion of two automatic procedures that at least partially implement this technique is given in \newcite{BE19rfla}. He showed that neural machine translations contain more strongly-associated formulaic sequences made of relatively high-frequency words, identified by the t-score, such as \textit{you know}, \textit{out of} or \textit{more than}, but far less strongly-associated formulaic sequences made of relatively rare words, identified by the MI, such as \textit{self-fulfilling prophecy}, \textit{sparsely populated} or \textit{sunnier climes}. 

These observations can be linked with a series of studies that have shown similar differences in foreign language learning \cite{BG14,Dur09} and which have proposed to interpret them in the framework of the usage-based model of language learning which "hold that a major determining force in the acquisition of formulas is the frequency of occurrence and co-occurrence of linguistic forms in the input" \cite{Dur09}. It is obviously tempting to use the same explanation for differences in translation, as neural models also seem to be affected by frequency of use \cite{Koehn17,Li2020}.

A competing explanation is however possible. All the texts analyzed in \newcite{TRITON2021} were quality newspaper articles written in French and published in \textit{Le Monde diplomatique}, and then translated in English for one of its international editions. However, as \newcite{Ponomarenko2019} pointed out following \newcite{Bielsa2009}, “Translation of news implies a higher degree of re-writing and re-telling than in any other type of translation” (p.40) and “International news, however, tends to prefer domestication of information instead of translation accuracy, which also has its particular reasons” (p.35). As it is important in this kind of texts that the translated version is as relevant and interesting as possible for the target readers, often from another culture, lexical and syntactic modifications, deletions and additions are frequent. All these modifications make the translation of this kind of texts less literal than the one expected from a machine translation system and thus risk to affect the differences in the use of formulaic sequences. 

\begin{figure*}
\centering
\includegraphics[width=.95\linewidth]{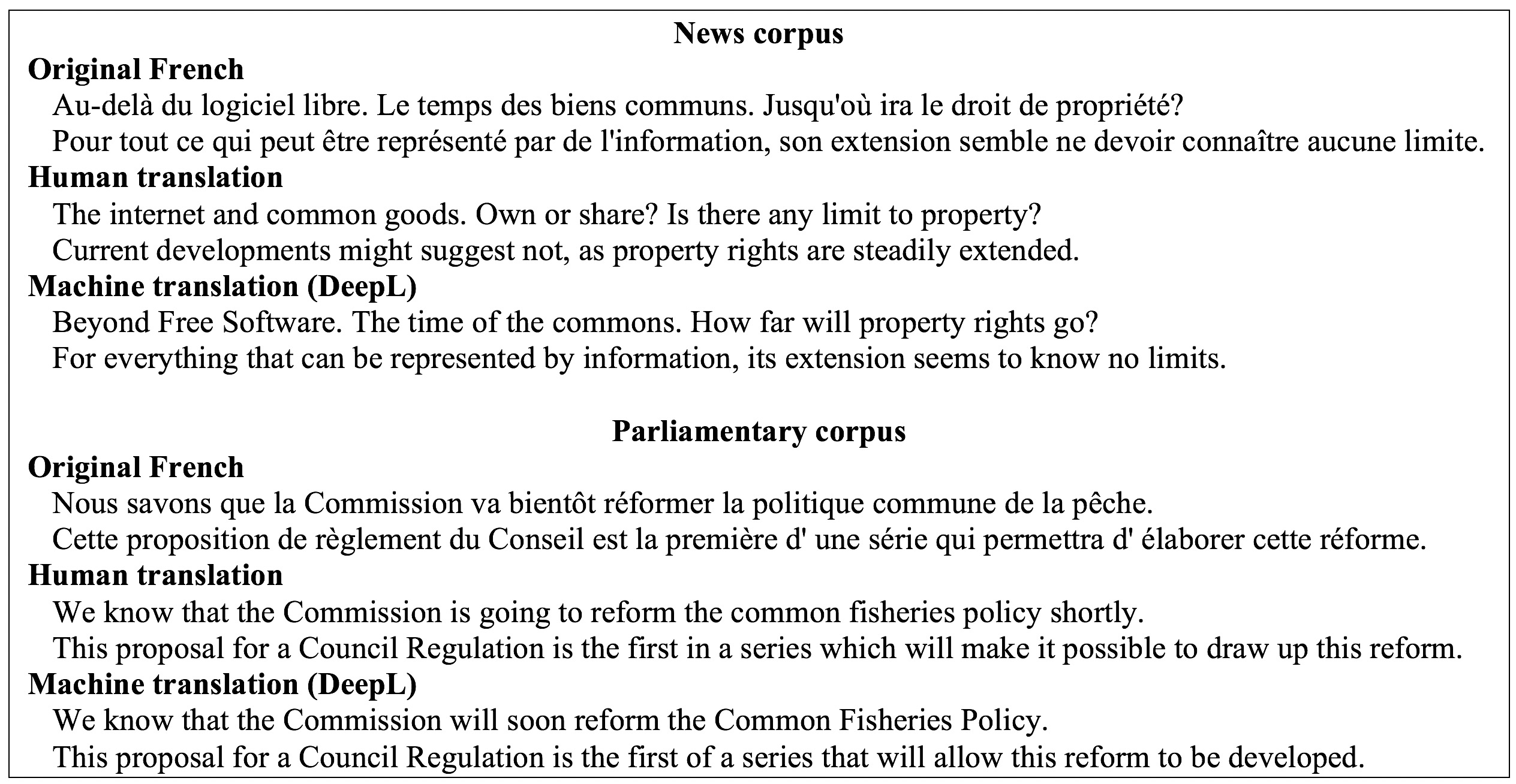}
\caption{Human and machine translation of a French excerpt from each of the two corpora.}
\end{figure*}

In this context, parliamentary corpora are a perfectly justified point of comparison. Translation accuracy is the main objective. For example, the criteria that European parliamentary debates translations must meet include: “the delivered target text is complete (no omissions nor additions are permitted)” and “the target text is a faithful, accurate and consistent translation of the source text” \cite{Sosoni2011}. 

Figure 1 illustrates this difference in translation between these two types of texts. It shows a brief extract from each corpus in its original and translated versions. The excerpt from the news corpus shows the different translation strategies used by the human and the machine, with the human version showing several reformulations affecting both the lexicon and the syntax and the deletion of one constituent in the last sentence. Such differences are not present in the extract from the parliamentary corpus. 

The objective of the present study is to determine whether machine translations of parliamentary texts differ from human translations in the use of phraseology, in order to confirm or refute the findings from the news corpus. The three following hypotheses are tested: compared to human translations, machine translations will contain more strongly-associated collocations made of high-frequency words, less strongly-associated formulaic sequences made of rare words and thus a larger ratio between these two indices. An positive conclusion, in addition to confirming the links between machine translation and foreign language learning, will suggest a way to make machine translations more similar to human translations, especially since the fully automatic nature of the analysis facilitates its large-scale use.

\section{Method}
\subsection{Parliamentary Corpus}
The material for this study is taken from the Europarl corpus v7 \cite{Koehn2005} available at \url{https://www.statmt.org/europarl}. In order to have parallel texts of which the original is in French and the translation in English, information rarely directly provided in the Europarl corpus because it was not developed for this purpose \cite{cartoni2012,Islam2012}, I employed the preprocessed version, freely available at \url{https://zenodo.org/record/1066474#.WnnEM3wiHcs}, obtained by means of the EuroparlExtract toolkit \cite{Ustaszewski2019}.
 
Two hundred texts of 3,500 to 4,500 characters, for a total of 120,000 words, were randomly selected among all texts written in French and translated into English. These thresholds were set in order to have texts long enough for collocation analysis, but not exceeding the 5,000-character limit imposed by the automatic translators used so that the document could be translated in a single operation.

Between February 14 and 16, 2022, three neural machine translation systems were used to translate these texts into English: the online version of \textit{DeepL} (\url{https://deepl.com/translator}) and \textit{Google Translate} (\url{https://translate.google.com}) and the \textit{Office 365} version of \textit{Microsoft Translator}.

\subsection{Procedure}

All contiguous word pairs (bigrams) were extracted from the CLAWS7 tokenized version \cite{Ray03} of each translated text. Bigrams, not including a proper noun, that could be found in the British National Corpus (BNC, \url{https://www.natcorp.ox.ac.uk}), were assigned an MI and a t-score on the basis of the frequencies in this reference corpus. Bigrams with $MI \geq 5$ or with $t \geq 6$ were deemed to be highly collocational \cite{ConfTrad,Dur09}. On this basis, three GollGram indices were calculated for each translated text: the percentage of highly collocational bigrams for MI, the same percentage for the t-score and the ratio between these two percentages (\%t-score $/$ \%MI). 

This procedure is in all points identical to the one used in the analysis of the news corpus \cite{TRITON2021}. The only difference is that the machine translations of the news corpus were undertaken in March-April 2021. According to \newcite{Porte2013} terminology, the present study is thus an \textit{approximate} replication that “might help us generalize, for example, the findings from the original study to a new population, setting, or modality” (p.11).

\begin{figure}
\centering
\includegraphics[width=.80\linewidth]{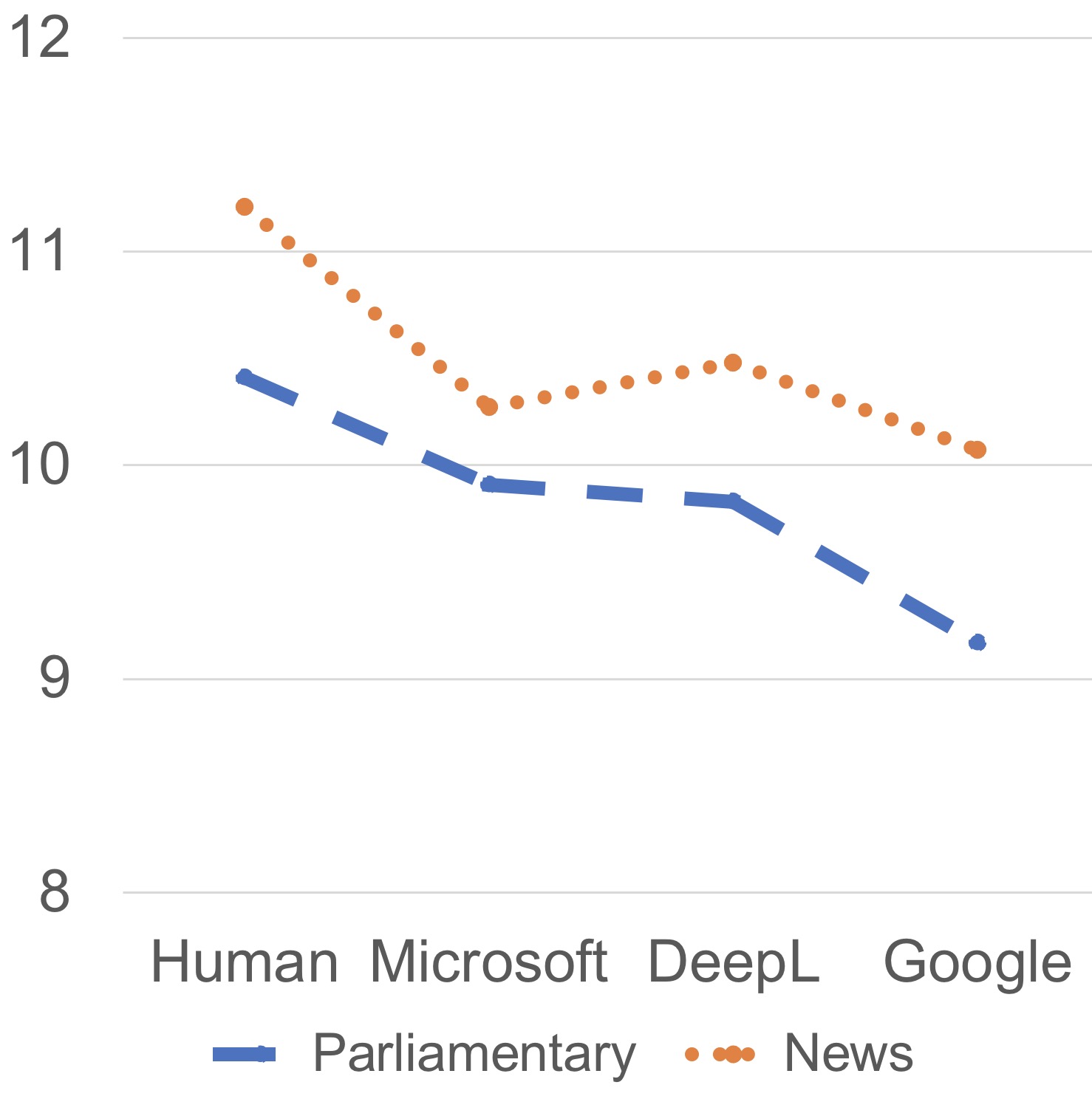}
\caption{Mean percentages of highly collocational bigrams for MI.}
\end{figure}

\begin{figure}
\centering
\includegraphics[width=.80\linewidth]{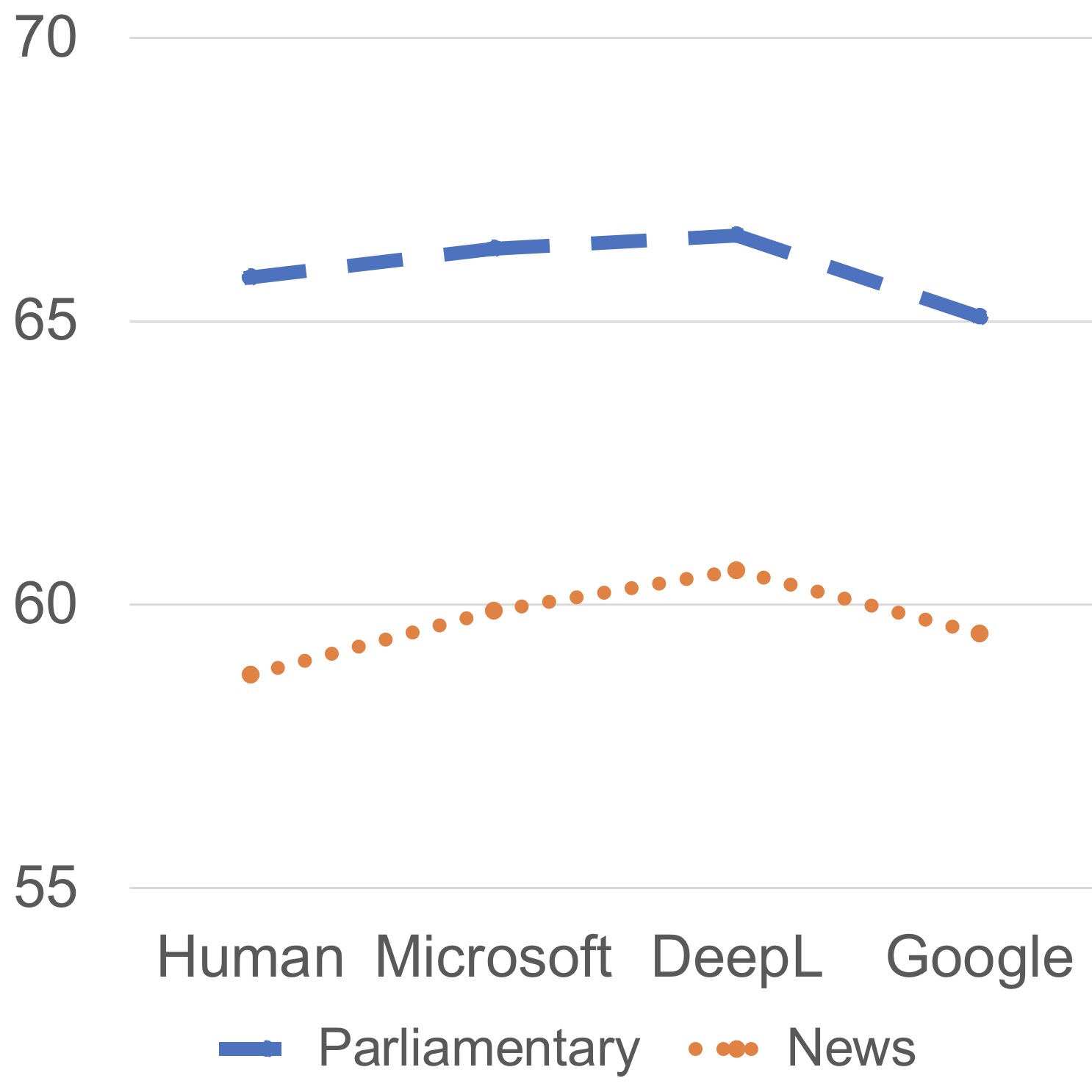}
\caption{Mean percentages of highly collocational bigrams for t-score.}
\end{figure}

\begin{figure}
\centering
\includegraphics[width=.80\linewidth]{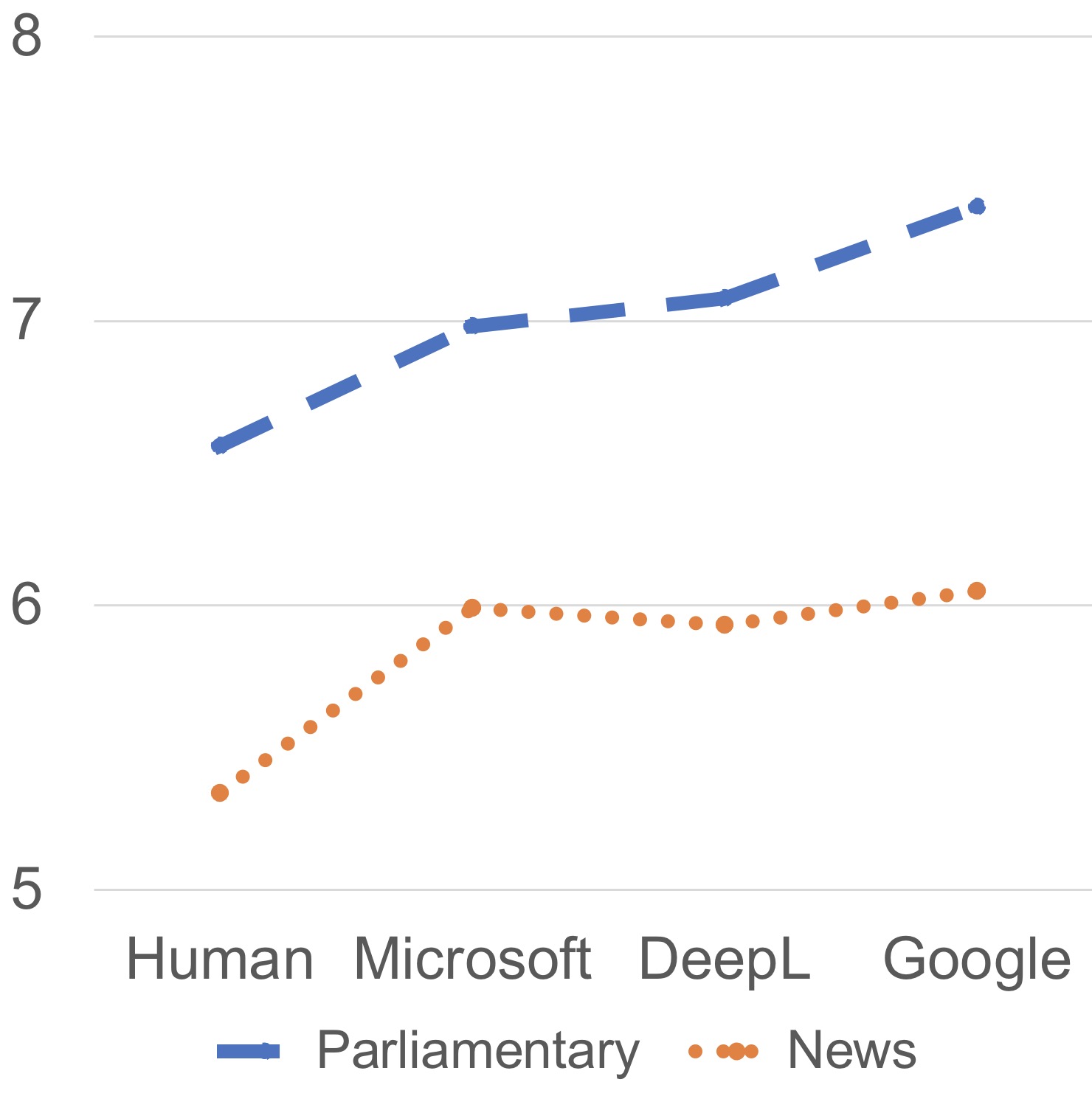}
\caption{Mean ratio between t-score and MI.}
\end{figure}

\begin{table}
\begin{center}
\begin{tabular}{rrrr}
\toprule
MI  &  Micro.     &   DeepL & Google  \\   \midrule
& \multicolumn{2}{c}{MI} & \\ 
Human \hspace{0.3cm} Di  & -0.50    & -0.59   &  -1.24 \\ 
   d & 0.38    & 0.44     & 0.89 \\ 
   p & 0.67    & 0.69     & 0.81 \\ 
Micro. \hspace{0.3cm} Di     &       & -0.09    & -0.74 \\ 
    d   &      & 0.11     & 0.67 \\ 
    p  &     & 0.55     & 0.76 \\ 
DeepL \hspace{0.3cm} Di        &  &            & -0.65 \\ 
    d      &  &           & 0.61 \\ 
    p       &  &          & 0.71 \\   \midrule
& \multicolumn{2}{c}{t-score} & \\ 
Human \hspace{0.3cm} Di   & 0.51    & 0.76     & -0.69 \\ 
   d & 0.15    & 0.31     & 0.26 \\ 
   p & 0.60    & 0.60     & 0.63 \\ 
Micro. \hspace{0.3cm} Di     &       & 0.25    & -1.20 \\ 
   s  &      & 0.09     & 0.45 \\ 
   p   &     & 0.55     & 0.80 \\ 
DeepL \hspace{0.3cm} Di     &  &              & -1.44 \\ 
   d   &  &              & 0.72 \\ 
   p   &  &             & 0.76 \\   \midrule
& \multicolumn{2}{c}{Ratio} & \\ 
Human \hspace{0.3cm} Di   & 0.42    & 0.52     & 0.84 \\ 
   d & 0.39    & 0.50     & 0.77 \\ 
   p & 0.69    & 0.70     & 0.81 \\ 
Micro. \hspace{0.3cm} Di   &         & 0.10     & 0.42 \\ 
   d  &        & 0.14     & 0.47 \\ 
   p   &       & 0.56     & 0.71 \\ 
DeepL \hspace{0.3cm} Di   &  &                 & 0.31 \\ 
   d  &  &                & 0.37 \\ 
   p   &  &               & 0.66 \\   
  \bottomrule
\end{tabular}
\caption{Differences (column translator minus row translator) and effect sizes for the two indices and the ratio in the four translation types. Di = Difference, d = Cohen's d, p = proportion of texts in which the mean effect is observed.}
\end{center}
\end{table}

\section{Results}
\subsection{Parliamentary Corpus}

The mean percentages of highly collocational bigrams for the MI and t-score and the mean ratio for the four translation type of the two genres of text are shown in Figures 2 to 4. The differences observed on the parliamentary corpus are very similar to those obtained with the news corpus. This section is focused on the analysis of the Parliamentary corpus while the comparison with the news corpus is presented in the next section. 

Table 1 presents the differences between the mean scores for every pairs of translators for the parliamentary corpus. The Student's t-test for non-independent measures was used to determine whether these mean differences were statistically significant. Due to the large number of tests performed (18), the Bonferroni procedure was used to compensate for the inflated Type I error rates, with the threshold for an alpha of 0.05 (two-tailed) set at 0.0027 (two-tailed). These tests indicate that all differences are significant, except for those between \textit{Microsoft Translator} and \textit{DeepL} for the three indices and the difference for the t-score between the human and \textit{Microsoft} translations. 

Table 1 also gives two effect sizes. Cohen's $d$ informs about the size of the difference between the means as a function of its variability. It is usual to consider that a $d$ of 0.20 indicates a small effect size, a $d$ of 0.50 a medium effect and a $d$ of 0.80 a large effect \cite{Cohen88}. The second effect size is the proportion of texts for which the difference between the two translations has the same sign as the mean difference. The maximum value of 1.0 means that texts produced by a translator always have larger scores than those translated by the other translator while the minimum value of 0.50 indicates no difference for this measure between the two translators.

As these results show, the three hypotheses about the differences between human and machine translations are all confirmed by statistically significant differences, except for the difference in t-score between human and Microsoft translations, which is nevertheless in the right direction. 

In these analyses, Cohen's $d$ for MI and for the ratio are often medium and sometimes even large and the differences are present in a large proportion of texts. For the t-score on the other hand, all effect sizes are small. This could be explained by the fact that the collocations highlighted by this lexical association measure are mostly very frequent in the language and thus more easily learned by automatic systems \cite{Koehn17,Li2020}. 

The comparison of the texts translated by \textit{Microsoft Translator} and by \textit{DeepL} does not show, as indicated above, any statistically significant difference and the effect sizes are very small. The outputs of \textit{Google Translate} on the other hand contain fewer highly collocational bigrams for MI and for t-score than these two other machine translators, potentially suggesting less efficiency of this translator for collocation processing.

\subsection{Comparison between the Two Genres}

As shown in Figures 2 to 4, all the observed trends are very similar in the two corpora, indicating that the two genres of text lead to the same conclusions. However, there is a strong contrast in the mean values. The MI scores are lower in the parliamentary corpus while the t-scores are higher. This is the case for both human and machine translations. This observation is most probably explained by a difference between the text genres, a difference that should already be present in the original French texts. 

The comparison of the effect sizes between the two types of translation (Table 2 in \newcite{TRITON2021}) clearly indicates that the differences are much stronger in the news corpus. This observation is consistent with the hypothesis of a greater literalness of the human translations in the parliamentary corpus.

\section{Conclusion}

The analyses carried out on the parliamentary corpus have made it possible to replicate the conclusions obtained with a news corpus. The observed trends are very similar, but the differences are less strong in the parliamentary corpus. This observation seems to confirm the usefulness of the parliamentary genre for the comparison of human and machine translation. Indeed, one can think that the less literal nature of the translations in news \cite{Ponomarenko2019,Sosoni2011} favors the identification of differences between the two types of translations. The differences observed in the parliamentary corpus thus seem to be more directly related to the translators effectiveness rather than to other factors.

Among the directions for future research, there is certainly the analysis of translations from English to French, but also between other languages. Here again, the Europarl corpus, as well as parliamentary debates in multilingual countries, allows for a great deal of experimentation. A potential difficulty is that the approach used requires a reference corpus in the target language. This problem does not seem too serious since it has been shown that freely available Wacky corpora \cite{BAR09} can be used without altering the results of the CollGram technique \cite{Bes16Tal}. Confirming these results in other languages, but also in other genres of texts, would allow to take advantage of them to try to improve machine translation systems.

A unanswered question is whether identical results would be obtained on the basis of a genre-specific reference corpus, which is very easy to obtain for European parliamentary debates. This corpus would be composed of documents originally produced in English. It would also be interesting to use other approaches to phraseology, especially more qualitative ones, to confirm the conclusions. Finally, the difference in mean values between the two text genres justifies an analysis of the original texts with the same technique.

\section{Acknowledgements}

The author is a Research Associate of the Fonds de la Recherche Scientifique (F.R.S-FNRS).

\section{Bibliographical References}\label{reference}

\bibliographystyle{lrec2022-bib}
\bibliography{SemEval-2021Task1}

\end{document}